\documentclass[10pt,twocolumn,letterpaper]{article}

\usepackage{cvpr}
\usepackage{times}
\usepackage{epsfig}
\usepackage{graphicx}
\usepackage{amsmath}
\usepackage{amssymb}

\usepackage[bookmarks=false,colorlinks=true,linkcolor=black,citecolor=black,filecolor=black,urlcolor=black]{hyperref}

\usepackage{url}
\usepackage{tabulary}

\usepackage{multirow}
\usepackage[normalem]{ulem}
\usepackage[british,UKenglish,USenglish,english,american]{babel}

\cvprfinalcopy % *** Uncomment this line for the final submission

\newcommand{\ve}[1]{\mathbf{#1}} % for displaying a vector
\newcommand{\tabincell}[2]{\begin{tabular}{@{}#1@{}}#2\end{tabular}}

\newcommand{\m}{$\times$}

\ifcvprfinal\fi
\begin{document}

%%%%%%%%% TITLE
\title{Aggregated Residual Transformations for Deep Neural Networks}

\author{
Saining Xie$^1$
\qquad
Ross Girshick$^2$
\qquad
Piotr Doll\'ar$^2$
\qquad
Zhuowen Tu$^1$
\qquad
Kaiming He$^2$
\\
$^1$UC San Diego
\qquad
$^2$Facebook AI Research\\
{\tt\small \{s9xie,ztu\}@ucsd.edu}
\qquad\quad
{\tt\small \{rbg,pdollar,kaiminghe\}@fb.com}
}

\maketitle
%\thispagestyle{empty}

%%%%%%%%% ABSTRACT
\begin{abstract}
\vspace{-.5em}
We present a simple, highly modularized network architecture for image classification.
Our network is constructed by repeating a building block that aggregates a set of transformations with the same topology.
Our simple design results in a homogeneous, multi-branch architecture that has only a few hyper-parameters to set. This strategy exposes a new dimension, which we call ``cardinality'' (the size of the set of transformations), as an essential factor in addition to the dimensions of depth and width. On the ImageNet-1K dataset, we empirically show that even under the restricted condition of maintaining complexity, increasing cardinality is able to improve classification accuracy. Moreover, increasing cardinality is more effective than going deeper or wider when we increase the capacity. Our models, named ResNeXt, are the foundations of our entry to the ILSVRC 2016 classification task in which we secured 2nd place.
We further investigate ResNeXt on an ImageNet-5K set and the COCO detection set, also showing better results than its ResNet counterpart. The code and models are publicly available online\footnote{https://github.com/facebookresearch/ResNeXt}.
\end{abstract}

%%%%%%%%% BODY TEXT
\section{Introduction}
\label{sec:intro}

Research on visual recognition is undergoing a transition from ``feature engineering'' to ``network engineering'' \cite{LeCun1989,Krizhevsky2012,Zeiler2014,Sermanet2014,Simonyan2015,Szegedy2015,He2016}.
In contrast to traditional hand-designed features (\eg, SIFT \cite{Lowe2004} and HOG \cite{Dalal2005}), features learned by neural networks from large-scale data \cite{Russakovsky2015} require minimal human involvement during training, and can be transferred to a variety of recognition tasks \cite{Donahue2014,Girshick2014,Long2015}.
Nevertheless, human effort has been shifted to designing better network architectures for learning representations.

Designing architectures becomes increasingly difficult with the growing number of hyper-parameters (width\footnote{Width refers to the number of channels in a layer.}, filter sizes, strides, \etc), especially when there are many layers.
The VGG-nets \cite{Simonyan2015} exhibit a simple yet effective strategy of constructing very deep networks: stacking building blocks of the same shape. This strategy is inherited by ResNets \cite{He2016} which stack modules of the same topology.
This simple rule reduces the free choices of hyper-parameters, and depth is exposed as an \emph{essential dimension} in neural networks.
Moreover, we argue that the simplicity of this rule may reduce the risk of over-adapting the hyper-parameters to a specific dataset. The robustness of VGG-nets and ResNets has been proven by various visual recognition tasks \cite{Donahue2014,Girshick2014,Girshick2015,Long2015,Pinheiro2015,He2016} and by non-visual tasks involving speech \cite{Xiong2016,Oord2016} and language \cite{Conneau2016,Wu2016,Kalchbrenner2016}.

\begin{figure}[t]
\centering
\includegraphics[width=1.0\linewidth]{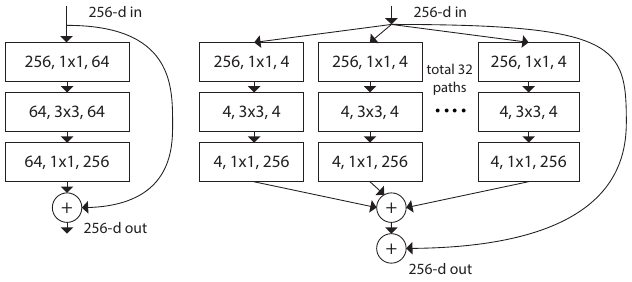}
\caption{\textbf{Left}: A block of ResNet \cite{He2016}. \textbf{Right}: A block of ResNeXt with cardinality $=32$, with roughly the same complexity. A layer is shown as (\# in channels, filter size, \# out channels).}
\label{fig:teaser}
\end{figure}

Unlike VGG-nets, the family of Inception models \cite{Szegedy2015,Ioffe2015,Szegedy2016a,Szegedy2016} have demonstrated that carefully designed topologies are able to achieve compelling accuracy with low theoretical complexity.
The Inception models have evolved over time \cite{Szegedy2015,Szegedy2016a}, but an important common property is a \emph{split-transform-merge} strategy. In an Inception module, the input is split into a few lower-dimensional embeddings (by 1$\times$1 convolutions), transformed by a set of specialized filters (3$\times$3, 5$\times$5, \etc), and merged by concatenation. 
It can be shown that the solution space of this architecture is a strict subspace of the solution space of a single large layer (\eg, 5$\times$5) operating on a high-dimensional embedding. The split-transform-merge behavior of Inception modules is expected to approach the representational power of large and dense layers, but at a considerably lower computational complexity.

Despite good accuracy, the realization of Inception models has been accompanied with a series of complicating factors --- the filter numbers and sizes are tailored for each individual transformation, and the modules are customized stage-by-stage. Although careful combinations of these components yield excellent neural network recipes, it is in general unclear how to adapt the Inception architectures to new datasets/tasks, especially when there are many factors and hyper-parameters to be designed.

In this paper, we present a simple architecture which adopts VGG/ResNets' strategy of repeating layers, while exploiting the split-transform-merge strategy in an easy, extensible way.
A module in our network performs a set of transformations, each on a low-dimensional embedding, whose outputs are aggregated by summation.
We pursuit a simple realization of this idea --- the transformations to be aggregated are all of the same topology (\eg, Fig.~\ref{fig:teaser} (right)). This design allows us to extend to any large number of transformations without specialized designs.

Interestingly, under this simplified situation we show that our model has two other equivalent forms (Fig.~\ref{fig:blocks}). The reformulation in Fig.~\ref{fig:blocks}(b) appears similar to the Inception-ResNet module \cite{Szegedy2016} in that it concatenates multiple paths; but our module differs from all existing Inception modules in that all our paths share the same topology and thus the number of paths can be easily isolated as a factor to be investigated. In a more succinct reformulation, our module can be reshaped by Krizhevsky \etal's grouped convolutions \cite{Krizhevsky2012} (Fig.~\ref{fig:blocks}(c)), which, however, had been developed as an engineering compromise.

We empirically demonstrate that our aggregated transformations outperform the original ResNet module, even under the restricted condition of maintaining computational complexity and model size --- \eg, Fig.~\ref{fig:teaser}(right) is designed to keep the FLOPs complexity and number of parameters of Fig.~\ref{fig:teaser}(left).
We emphasize that while it is relatively easy to increase accuracy by increasing capacity (going deeper or wider), methods that increase accuracy while maintaining (or reducing) complexity are rare in the literature.

Our method indicates that \emph{cardinality} (the size of the set of transformations) is a concrete, measurable dimension that is of central importance, in addition to the dimensions of width and depth.
Experiments demonstrate that \emph{increasing cardinality is a more effective way of gaining accuracy than going deeper or wider}, especially when depth and width starts to give diminishing returns for existing models.

Our neural networks, named \emph{ResNeXt} (suggesting the \emph{next} dimension), outperform ResNet-101/152 \cite{He2016}, ResNet-200 \cite{He2016a}, Inception-v3 \cite{Szegedy2016a}, and Inception-ResNet-v2 \cite{Szegedy2016} on the ImageNet classification dataset. 
In particular, a 101-layer ResNeXt is able to achieve better accuracy than ResNet-200 \cite{He2016a} but has only 50\% complexity. Moreover, ResNeXt exhibits considerably simpler designs than all Inception models. 
ResNeXt was the foundation of our submission to the ILSVRC 2016 classification task, in which we secured second place.
This paper further evaluates ResNeXt on a larger ImageNet-5K set and the COCO object detection dataset \cite{Lin2014}, showing consistently better accuracy than its ResNet counterparts.
We expect that ResNeXt will also generalize well to other visual (and non-visual) recognition tasks.

\section{Related Work}

\noindent\textbf{Multi-branch convolutional networks}. The Inception models \cite{Szegedy2015,Ioffe2015,Szegedy2016a,Szegedy2016} are successful multi-branch architectures where each branch is carefully customized.
ResNets \cite{He2016} can be thought of as two-branch networks where one branch is the identity mapping.
Deep neural decision forests \cite{Kontschieder2015} are tree-patterned multi-branch networks with learned splitting functions.

\vspace{.5em}
\noindent\textbf{Grouped convolutions}. The use of grouped convolutions dates back to the AlexNet paper \cite{Krizhevsky2012}, if not earlier. The motivation given by Krizhevsky \etal \cite{Krizhevsky2012} is for distributing the model over two GPUs. Grouped convolutions are supported by Caffe \cite{Jia2014}, Torch \cite{Collobert2002}, and other libraries, mainly for compatibility of AlexNet. To the best of our knowledge, there has been little evidence on exploiting grouped convolutions to \emph{improve} accuracy.
A special case of grouped convolutions is channel-wise convolutions in which the number of groups is equal to the number of channels. Channel-wise convolutions are part of the separable convolutions in \cite{Sifre2014}.

\vspace{.5em}
\noindent\textbf{Compressing convolutional networks}.
Decomposition (at spatial \cite{Denton2014,Jaderberg2014} and/or channel \cite{Denton2014,Kim2016,Ioannou2016} level) is a widely adopted technique to reduce redundancy of deep convolutional networks and accelerate/compress them. Ioannou \etal \cite{Ioannou2016} present a ``root''-patterned network for reducing computation, and branches in the root are realized by grouped convolutions. These methods \cite{Denton2014,Jaderberg2014,Kim2016,Ioannou2016} have shown elegant compromise of accuracy with lower complexity and smaller model sizes. Instead of compression, our method is an architecture that empirically shows stronger representational power.

\vspace{.5em}
\noindent\textbf{Ensembling}. 
Averaging a set of independently trained networks is an effective solution to improving accuracy \cite{Krizhevsky2012}, widely adopted in recognition competitions \cite{Russakovsky2015}. Veit \etal \cite{Veit2016} interpret a single ResNet as an ensemble of shallower networks, which results from ResNet's \emph{additive} behaviors \cite{He2016a}. Our method harnesses additions to aggregate a set of transformations. But we argue that it is imprecise to view our method as ensembling, because the members to be aggregated are trained jointly, not independently.

\section{Method}

\subsection{Template}
\label{sec:template}

We adopt a highly modularized design following VGG/ResNets. Our network consists of a stack of residual blocks. These blocks have the same topology, and are subject to two simple rules inspired by VGG/ResNets: (i) if producing spatial maps of the same size, the blocks share the same hyper-parameters (width and filter sizes), and (ii) each time when the spatial map is downsampled by a factor of 2, the width of the blocks is multiplied by a factor of 2. The second rule ensures that the computational complexity, in terms of FLOPs (floating-point operations, in \# of multiply-adds), is roughly the same for all blocks.

With these two rules, we only need to design a \emph{template} module, and all modules in a network can be determined accordingly. So these two rules greatly narrow down the design space and allow us to focus on a few key factors.
The networks constructed by these rules are in Table~\ref{tab:arch}.

\newcommand{\blockb}[3]{\multirow{3}{*}{
\(\left[
\begin{array}{l}
\text{1$\times$1, #2}\\
[-.1em] \text{3$\times$3, #2}\\
[-.1em] \text{1$\times$1, #1}
\end{array}\right]\)$\times$#3}
}

\newcommand{\blockx}[3]{\multirow{3}{*}{
\(\left[
\begin{array}{l}
\text{1$\times$1, #2}\\
[-.1em] \text{3$\times$3, #2, $C$=32}\\
[-.1em] \text{1$\times$1, #1}\\
\end{array}\right]\)$\times$#3}
}

\newcolumntype{x}[1]{>\centering p{#1pt}}
\newcommand{\ft}[1]{\fontsize{#1pt}{1em}\selectfont}
\renewcommand\arraystretch{1.25}
\setlength{\tabcolsep}{1.2pt}
\begin{table}[t]
\begin{center}
\footnotesize
\begin{tabular}{c|c|x{80}|c}
\hline
 stage & output & ResNet-50 & \textbf{ResNeXt-50 (32\m4d)} \\
\hline
conv1 & \ft{7} 112$\times$112 & 7$\times$7, 64, stride 2 & 7$\times$7, 64, stride 2 \\
\hline
\multirow{4}{*}{conv2} & \multirow{4}{*}{\ft{7} 56$\times$56} & 3$\times$3 max pool, stride 2 & 3$\times$3 max pool, stride 2 \\\cline{3-4}
  &  &  \blockb{256}{64}{3} & \blockx{256}{128}{3}\\
  &  &  & \\
  &  &  & \\
\hline
\multirow{3}{*}{conv3} &  \multirow{3}{*}{\ft{7} 28$\times$28} 
  & \blockb{512}{128}{4} &  \blockx{512}{256}{4}\\
  &  &  & \\
  &  &  & \\
\hline
\multirow{3}{*}{conv4} & \multirow{3}{*}{\ft{7} 14$\times$14} 
  & \blockb{1024}{256}{6} & \blockx{1024}{512}{6}\\
  &  &  & \\
  &  &  & \\
\hline
\multirow{3}{*}{conv5} & \multirow{3}{*}{\ft{7} 7$\times$7} 
& \blockb{2048}{512}{3} & \blockx{2048}{1024}{3}\\
  &  &  & \\
  &  &  & \\
\hline
& \multirow{2}{*}{\ft{7} 1$\times$1} & global average pool & global average pool \\
 & & 1000-d fc, softmax & 1000-d fc, softmax \\
\hline
\multicolumn{2}{c|}{\small \# params.} & \small \textbf{25.5}$\times$$10^6$  & \small \textbf{25.0}$\times$$10^6$ \\
\hline
\multicolumn{2}{c|}{\small FLOPs} & \small \textbf{4.1}$\times$$10^9$  & \small \textbf{4.2}$\times$$10^9$ \\
\hline
\end{tabular}
\end{center}
%\vspace{1em}
\caption{(\textbf{Left}) ResNet-50. (\textbf{Right}) ResNeXt-50 with a 32\m4d template (using the reformulation in Fig.~\ref{fig:blocks}(c)).
Inside the brackets are the shape of a residual block, and outside the brackets is the number of stacked blocks on a stage. ``$C$=32'' suggests grouped convolutions \cite{Krizhevsky2012} with 32 groups. \emph{The numbers of parameters and FLOPs are similar between these two models.}
}
\label{tab:arch}
\vspace{-.5em}
\end{table}

\subsection{Revisiting Simple Neurons}

\begin{figure}[t]
\centering
\vspace{-1em}
\includegraphics[width=0.7\linewidth]{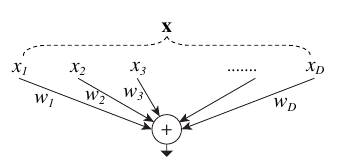}
\caption{A simple neuron that performs inner product.}
\label{fig:neuron}
\end{figure}

The simplest neurons in artificial neural networks perform inner product (weighted sum), which is the elementary transformation done by fully-connected and convolutional layers. 
Inner product can be thought of as a form of aggregating transformation:
\vspace{-.5em}
\begin{equation}\label{eq:inner}
\sum_{i=1}^{D}w_i x_i,
\vspace{-.5em}
\end{equation}
where $\ve{x}=[x_1, x_2, ..., x_{D}]$ is a $D$-channel input vector to the neuron and $w_i$ is a filter's weight for the $i$-th channel.
This operation (usually including some output non-linearity) is referred to as a ``neuron''. See Fig.~\ref{fig:neuron}.

The above operation can be recast as a combination of \emph{splitting, transforming, and aggregating}. (i) \emph{Splitting}: the vector $\ve{x}$ is sliced as a low-dimensional embedding, and in the above, it is a single-dimension subspace $x_i$. (ii) \emph{Transforming}: the low-dimensional representation is transformed, and in the above, it is simply scaled: $w_i x_i$.
(iii) \emph{Aggregating}: the transformations in all embeddings are aggregated by $\sum_{i=1}^{D}$.

\begin{figure*}[t]
\centering
\vspace{-1em}
\includegraphics[width=0.95\linewidth]{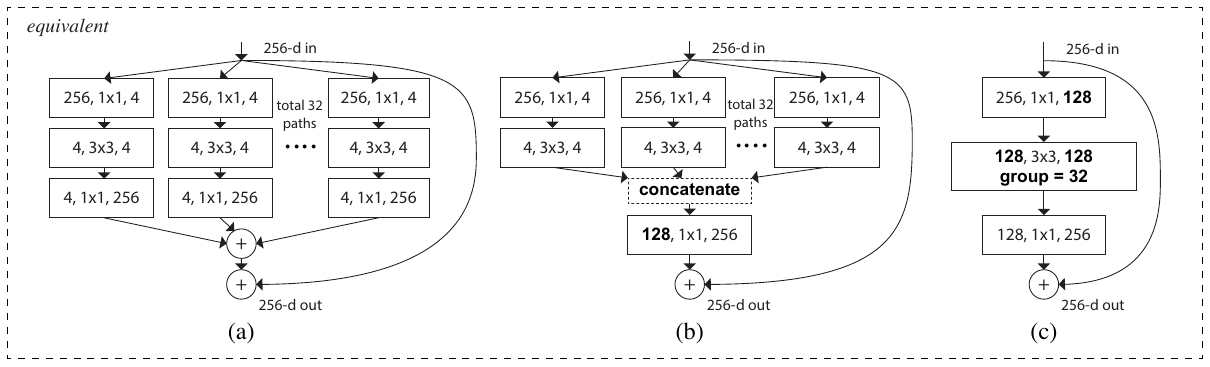}
\caption{Equivalent building blocks of ResNeXt.
\textbf{(a)}: Aggregated residual transformations, the same as Fig.~\ref{fig:teaser} right.
\textbf{(b)}: A block equivalent to (a), implemented as early concatenation.
\textbf{(c)}: A block equivalent to (a,b), implemented as grouped convolutions \cite{Krizhevsky2012}. Notations in \textbf{bold} text highlight the reformulation changes. A layer is denoted as (\# input channels, filter size, \# output channels).
}
\label{fig:blocks}
\end{figure*}

\subsection{Aggregated Transformations}

Given the above analysis of a simple neuron, we consider replacing the elementary transformation ($w_i x_i$) with a more generic function, which in itself can also be a network. In contrast to ``Network-in-Network" \cite{Lin2014a} that turns out to increase the dimension of depth, we show that our ``Network-in-\emph{Neuron}'' expands along a new dimension.

Formally, we present aggregated transformations as:
\vspace{-.5em}
\begin{equation}\label{eq:general}
\mathcal{F}(\ve{x})=\sum_{i=1}^{C}\mathcal{T}_i(\ve{x}),
\vspace{-.5em}
\end{equation}
where $\mathcal{T}_i(\ve{x})$ can be an arbitrary function. Analogous to a simple neuron, $\mathcal{T}_i$ should project $\ve{x}$ into an (optionally low-dimensional) embedding and then transform it. 

In Eqn.(\ref{eq:general}), $C$ is the size of the set of transformations to be aggregated. We refer to $C$ as \emph{cardinality} \cite{Cantor1884}.
In Eqn.(\ref{eq:general}) $C$ is in a position similar to $D$ in Eqn.(\ref{eq:inner}), but $C$ need not equal $D$ and can be an arbitrary number. While the dimension of width is related to the number of simple transformations (inner product), we argue that the dimension of cardinality controls the number of more complex transformations. We show by experiments that cardinality is an essential dimension and can be more effective than the dimensions of width and depth.

In this paper, we consider a simple way of designing the transformation functions: all $\mathcal{T}_i$'s have the same topology. This extends the VGG-style strategy of repeating layers of the same shape, which is helpful for isolating a few factors and extending to any large number of transformations. We set the individual transformation $\mathcal{T}_i$ to be the bottleneck-shaped architecture \cite{He2016}, as illustrated in Fig.~\ref{fig:teaser} (right). In this case, the first 1$\times$1 layer in each $\mathcal{T}_i$ produces the low-dimensional embedding.

The aggregated transformation in Eqn.(\ref{eq:general}) serves as the residual function \cite{He2016} (Fig.~\ref{fig:teaser} right):
\vspace{-.5em}
\begin{equation}\label{eq:resnext}
\ve{y}=\ve{x}+\sum_{i=1}^{C}\mathcal{T}_i(\ve{x}),
\vspace{-.5em}
\end{equation}
where $\ve{y}$ is the output.

\vspace{.5em}
\noindent\emph{Relation to Inception-ResNet}. Some tensor manipulations show that the module in Fig.~\ref{fig:teaser}(right) (also shown in Fig.~\ref{fig:blocks}(a)) is equivalent to Fig.~\ref{fig:blocks}(b).\footnote{An informal but descriptive proof is as follows. Note the equality: $A_1B_1+A_2B_2 = [A_1, A_2] [B_1; B_2]
$ where $[~, ~]$ is horizontal concatenation and $[~; ~]$ is vertical concatenation. Let $A_i$ be the weight of the last layer and $B_i$ be the output response of the second-last layer in the block. In the case of $C=2$, the element-wise addition in Fig.~\ref{fig:blocks}(a) is $A_1B_1+A_2B_2$, the weight of the last layer in Fig.~\ref{fig:blocks}(b) is $[A_1, A_2]$, and the concatenation of outputs of second-last layers in Fig.~\ref{fig:blocks}(b) is $[B_1; B_2]$.}
Fig.~\ref{fig:blocks}(b) appears similar to the Inception-ResNet \cite{Szegedy2016} block in that it involves branching and concatenating in the residual function. But unlike all Inception or Inception-ResNet modules, we share the same topology among the multiple paths. Our module requires minimal extra effort designing each path.

\vspace{.5em}
\noindent\emph{Relation to Grouped Convolutions}. The above module becomes more succinct using the notation of \emph{grouped convolutions} \cite{Krizhevsky2012}.\footnote{In a group conv layer \cite{Krizhevsky2012}, input and output channels are divided into $C$ groups, and convolutions are separately performed within each group.} This reformulation is illustrated in Fig.~\ref{fig:blocks}(c).
All the low-dimensional embeddings (the first 1$\times$1 layers) can be replaced by a single, wider layer (\eg, 1$\times$1, 128-d in Fig~\ref{fig:blocks}(c)). 
Splitting is essentially done by the grouped convolutional layer when it divides its input channels into groups.
The grouped convolutional layer in Fig.~\ref{fig:blocks}(c) performs 32 groups of convolutions whose input and output channels are 4-dimensional. The grouped convolutional layer concatenates them as the outputs of the layer.
The block in Fig.~\ref{fig:blocks}(c) looks like the original bottleneck residual block in Fig.~\ref{fig:teaser}(left), except that Fig.~\ref{fig:blocks}(c) is a wider but sparsely connected module.

\begin{figure}[t]
\centering
\includegraphics[width=1.0\linewidth]{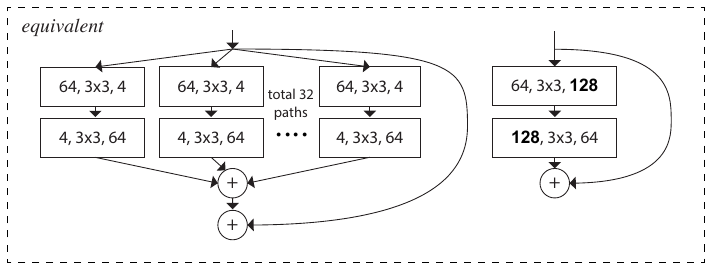}
\caption{(\textbf{Left}): Aggregating transformations of depth $= 2$. (\textbf{Right}): An equivalent block, which is trivially wider.}
\label{fig:2layer}
\end{figure}

We note that the reformulations produce nontrivial topologies only when the block has depth $\geq$3. If the block has depth $=$ 2 (\eg, the basic block in \cite{He2016}), the reformulations lead to trivially a wide, dense module. See the illustration in Fig.~\ref{fig:2layer}.

\vspace{.5em}
\noindent\emph{Discussion}. We note that although we present reformulations that exhibit concatenation (Fig.~\ref{fig:blocks}(b)) or grouped convolutions (Fig.~\ref{fig:blocks}(c)), such reformulations are not always applicable for the general form of Eqn.(\ref{eq:resnext}), \eg, if the transformation $\mathcal{T}_i$ takes arbitrary forms and are heterogenous.
We choose to use homogenous forms in this paper because they are simpler and extensible. Under this simplified case, grouped convolutions in the form of Fig.~\ref{fig:blocks}(c) are helpful for easing implementation.

\subsection{Model Capacity}
\label{sec:capacity}

Our experiments in the next section will show that our models improve accuracy when maintaining the model complexity and number of parameters. This is not only interesting in practice, but more importantly, the complexity and number of parameters represent inherent capacity of models and thus are often investigated as fundamental properties of deep networks \cite{Eigen2013a}.

When we evaluate different cardinalities $C$ while preserving complexity, we want to minimize the modification of other hyper-parameters.
We choose to adjust the width of the bottleneck (\eg, 4-d in Fig~\ref{fig:teaser}(right)), because it can be isolated from the input and output of the block. This strategy introduces no change to other hyper-parameters (depth or input/output width of blocks), so is helpful for us to focus on the impact of cardinality.

In Fig.~\ref{fig:teaser}(left), the original ResNet bottleneck block \cite{He2016} has $256\cdot64+3\cdot3\cdot64\cdot64+64\cdot256\approx $ 70k parameters and proportional FLOPs (on the same feature map size). With bottleneck width $d$, our template in Fig.~\ref{fig:teaser}(right) has:
%\vspace{-.5em}
\begin{equation}\label{eq:capacity}
C \cdot (256\cdot d + 3\cdot 3 \cdot d\cdot d + d \cdot 256)
%\vspace{-.5em}
\end{equation}
parameters and proportional FLOPs.
When $C=32$ and $d=4$, Eqn.(\ref{eq:capacity}) $\approx$ 70k. Table~\ref{tab:capacity} shows the relationship between cardinality $C$ and bottleneck width $d$.

Because we adopt the two rules in Sec.~\ref{sec:template}, the above approximate equality is valid between a ResNet bottleneck block and our ResNeXt on all stages (except for the subsampling layers where the feature maps size changes).
Table~\ref{tab:arch} compares the original ResNet-50 and our ResNeXt-50 that is of similar capacity.\footnote{The marginally smaller number of parameters and marginally higher FLOPs are mainly caused by the blocks where the map sizes change.}
We note that the complexity can only be preserved approximately, but the difference of the complexity is minor and does not bias our results.

\renewcommand\arraystretch{1.0}
\setlength{\tabcolsep}{6pt}
\begin{table}[t]
\begin{center}
\footnotesize
\begin{tabular}{c|ccccc}
\hline
cardinality $C$ & 1 & 2 & 4 & 8 & 32 \\
\hline
width of bottleneck $d$ & 64 & 40 & 24 & 14 & 4 \\
width of group conv. & 64 & 80 & 96 & 112 & 128 \\
\hline
\end{tabular}
\end{center}
\caption{Relations between cardinality and width (for the template of conv2), with roughly preserved complexity on a residual block. The number of parameters is $\sim$70k for the template of conv2. The number of FLOPs is $\sim$0.22 billion (\# params$\times$56$\times$56 for conv2).
}
\label{tab:capacity}
\end{table}

\begin{figure*}[t]
\centering
\vspace{-2em}
\includegraphics[width=.48\linewidth]{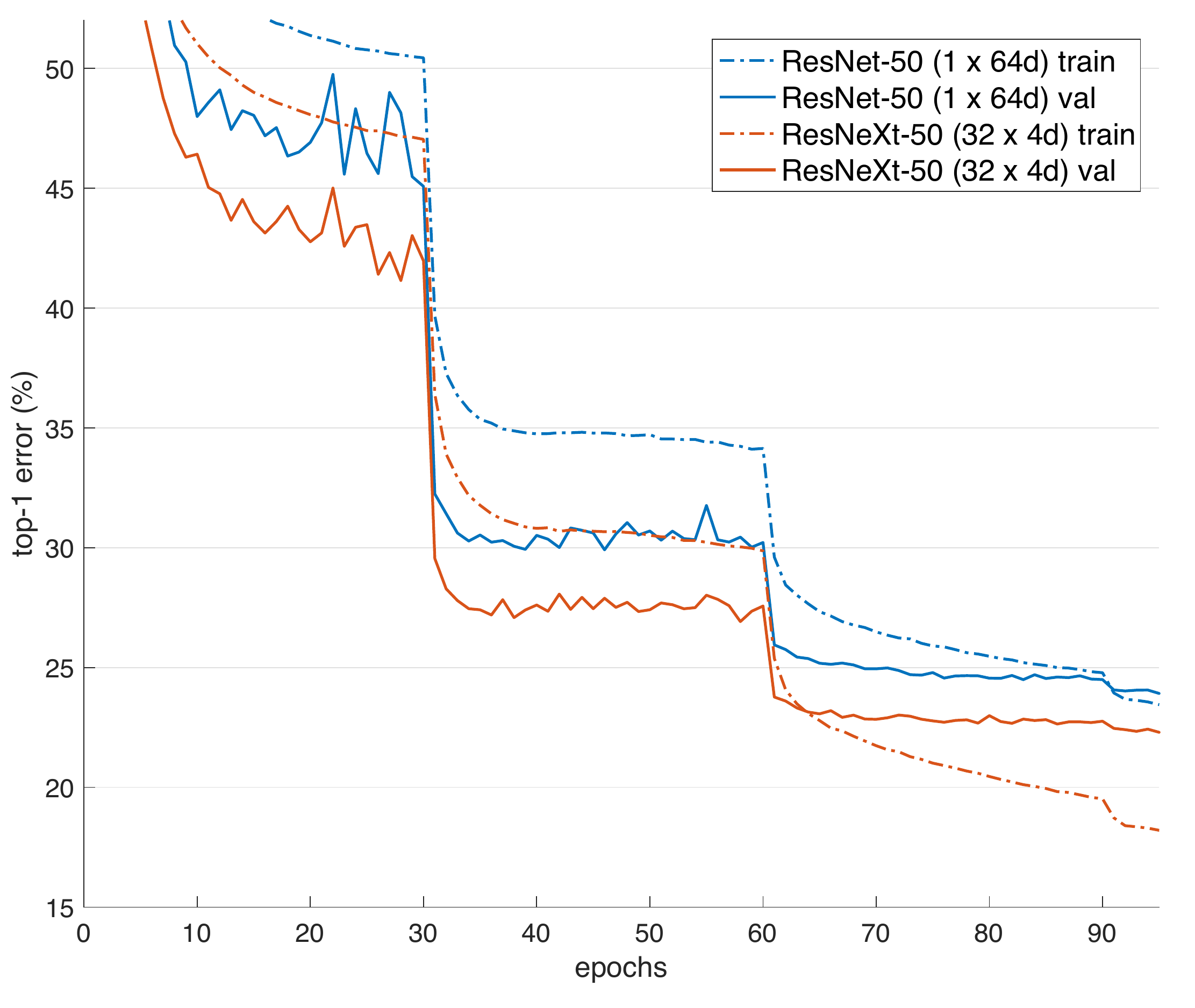}
\includegraphics[width=.48\linewidth]{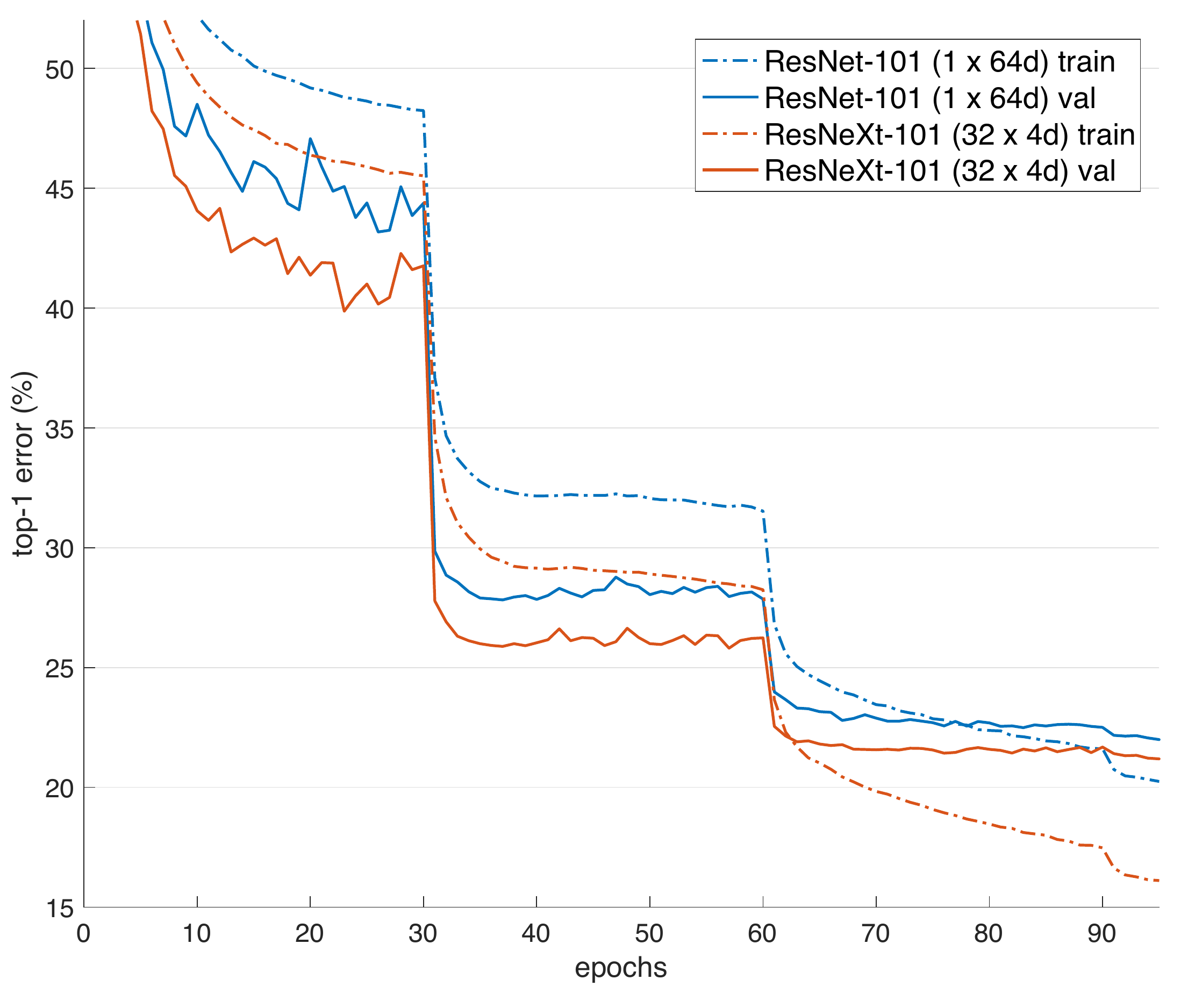}
\caption{Training curves on ImageNet-1K. (\textbf{Left}): ResNet/ResNeXt-50 with preserved complexity ($\sim$4.1 billion FLOPs, $\sim$25 million parameters); (\textbf{Right}): ResNet/ResNeXt-101 with preserved complexity ($\sim$7.8 billion FLOPs, $\sim$44 million parameters).
}
\label{fig:curves-comp}
\end{figure*}

\section{Implementation details}
\label{sec:impl}

Our implementation follows \cite{He2016} and the publicly available code of \texttt{fb.resnet.torch} \cite{Gross2016}. On the ImageNet dataset, the input image is 224$\times$224 randomly cropped from a resized image using the scale and aspect ratio augmentation of \cite{Szegedy2015} implemented by \cite{Gross2016}. 
The shortcuts are identity connections except for those increasing dimensions which are projections (type B in \cite{He2016}).
Downsampling of conv3, 4, and 5 is done by stride-2 convolutions in the 3$\times$3 layer of the first block in each stage, as suggested in \cite{Gross2016}.
We use SGD with a mini-batch size of 256 on 8 GPUs (32 per GPU).
The weight decay is 0.0001 and the momentum is 0.9. We start from a learning rate of 0.1, and divide it by 10 for three times using the schedule in \cite{Gross2016}. We adopt the weight initialization of \cite{He2015}. In all ablation comparisons, we evaluate the error on the single 224$\times$224 center crop from an image whose shorter side is 256.

Our models are realized by the form of Fig.~\ref{fig:blocks}(c). 
We perform batch normalization (BN) \cite{Ioffe2015} right after the convolutions in Fig.~\ref{fig:blocks}(c).\footnote{With BN, for the equivalent form in Fig.~\ref{fig:blocks}(a), BN is employed after aggregating the transformations and before adding to the shortcut.} ReLU is performed right after each BN, expect for the output of the block where ReLU is performed after the adding to the shortcut, following \cite{He2016}.

We note that the three forms in Fig.~\ref{fig:blocks} are strictly equivalent, when BN and ReLU are appropriately addressed as mentioned above. We have trained all three forms and obtained the same results. We choose to implement by Fig.~\ref{fig:blocks}(c) because it is more succinct and faster than the other two forms.

\section{Experiments}

\subsection{Experiments on ImageNet-1K}

We conduct ablation experiments on the 1000-class ImageNet classification task \cite{Russakovsky2015}. We follow \cite{He2016} to construct 50-layer and 101-layer residual networks. We simply replace all blocks in ResNet-50/101 with our blocks.

\vspace{.5em}
\noindent\textbf{Notations}.
Because we adopt the two rules in Sec.~\ref{sec:template}, it is sufficient for us to refer to an architecture by the template.
For example, Table~\ref{tab:arch} shows a ResNeXt-50 constructed by a template with cardinality $= 32$ and bottleneck width $=$ 4d (Fig.~\ref{fig:blocks}). This network is denoted as ResNeXt-50 (\textbf{32\m4d}) for simplicity. We note that the input/output width of the template is fixed as 256-d (Fig.~\ref{fig:blocks}), and all widths are doubled each time when the feature map is subsampled (see Table~\ref{tab:arch}).

\vspace{.5em}
\noindent\textbf{Cardinality \vs Width}.
We first evaluate the trade-off between cardinality $C$ and bottleneck width, under preserved complexity as listed in Table~\ref{tab:capacity}.
Table~\ref{tab:preserve-comp} shows the results and Fig.~\ref{fig:curves-comp} shows the curves of error \vs epochs.
Comparing with ResNet-50 (Table~\ref{tab:preserve-comp} top and Fig.~\ref{fig:curves-comp} left), the 32\m4d ResNeXt-50 has a validation error of 22.2\%, which is \textbf{1.7\%} lower than the ResNet baseline's 23.9\%. With cardinality $C$ increasing from 1 to 32 while keeping complexity, the error rate keeps reducing.
Furthermore, the 32\m4d ResNeXt also has a much lower \emph{training error} than the ResNet counterpart, suggesting that the gains are \emph{not} from regularization but from \emph{stronger representations}.

Similar trends are observed in the case of ResNet-101 (Fig.~\ref{fig:curves-comp} right, Table~\ref{tab:preserve-comp} bottom), where the 32\m4d ResNeXt-101 outperforms the ResNet-101 counterpart by 0.8\%. Although this improvement of validation error is smaller than that of the 50-layer case, the improvement of training error is still big (20\% for ResNet-101 and 16\% for 32\m4d ResNeXt-101, Fig.~\ref{fig:curves-comp} right). 
In fact, more training data will enlarge the gap of validation error, as we show on an ImageNet-5K set in the next subsection.

Table~\ref{tab:preserve-comp} also suggests that \emph{with complexity preserved}, increasing cardinality at the price of reducing width starts to show saturating accuracy when the bottleneck width is small. We argue that it is not worthwhile to keep reducing width in such a trade-off. So we adopt a bottleneck width no smaller than 4d in the following.

\renewcommand\arraystretch{1.1}
\setlength{\tabcolsep}{12pt}
\begin{table}[t]
\begin{center}
\small
\begin{tabular}{l|c|c}
\hline
	& setting & \footnotesize top-1 error (\%) \\
\hline
ResNet-50		& 1 \m~64d		& 23.9 \\
ResNeXt-50		& 2 \m~40d		& 23.0 \\
ResNeXt-50		& 4 \m~24d		& 22.6 \\
ResNeXt-50		& 8 \m~14d		& 22.3 \\
ResNeXt-50		& 32 \m~4d		& \textbf{22.2} \\
\hline
ResNet-101		& 1 \m~64d		& 22.0 \\
ResNeXt-101	& 2 \m~40d		& 21.7 \\
ResNeXt-101	& 4 \m~24d		& 21.4 \\
ResNeXt-101	& 8 \m~14d		& 21.3 \\
ResNeXt-101	& 32 \m~4d		& \textbf{21.2} \\
\hline
\end{tabular}
\end{center}
\caption{Ablation experiments on ImageNet-1K. (\textbf{Top}): ResNet-50 with preserved complexity ($\sim$4.1 billion FLOPs); (\textbf{Bottom}): ResNet-101 with preserved complexity ($\sim$7.8 billion FLOPs). The error rate is evaluated on the single crop of 224$\times$224 pixels.}
\label{tab:preserve-comp}
\end{table}

\renewcommand\arraystretch{1.1}
\setlength{\tabcolsep}{4pt}
\begin{table}[t]
\begin{center}
\small
\begin{tabular}{l|c|c|c}
\hline
	& \footnotesize setting
	& \footnotesize  top-1 err (\%)
	& \footnotesize  top-5 err (\%) \\
\hline
\multicolumn{4}{l}{\emph{1$\times$ complexity references:}} \\
\hline
ResNet-101												& 1 \m~64d					& 22.0 				& 6.0 \\
ResNeXt-101												& 32 \m~4d					& 21.2 				& 5.6 \\
\hline
\multicolumn{4}{l}{\emph{2$\times$ complexity models follow:}} \\
\hline
ResNet-\textbf{200} \cite{He2016a}		& 1 \m~64d					& 21.7				& 5.8 \\
ResNet-101, wider									& 1 \m~\textbf{100}d	& 21.3 				& 5.7 \\
\hline
ResNeXt-101								& \textbf{2} \m~64d					& 20.7	& 5.5 \\
ResNeXt-101								& \textbf{64}	 \m~4d					& \textbf{20.4}	& \textbf{5.3} \\
\hline
\end{tabular}
\end{center}
\caption{Comparisons on ImageNet-1K when the number of FLOPs is increased to 2$\times$ of ResNet-101's.
The error rate is evaluated on the single crop of 224$\times$224 pixels. The highlighted factors are the factors that increase complexity.}
\label{tab:2x-complexity}
\end{table}

\vspace{.5em}
\noindent\textbf{Increasing Cardinality \vs Deeper/Wider}.
Next we investigate increasing complexity by increasing cardinality $C$ or increasing depth or width. The following comparison can also be viewed as with reference to 2$\times$ FLOPs of the ResNet-101 baseline. We compare the following variants that have $\sim$15 billion FLOPs.
(i) \textbf{Going deeper} to 200 layers. We adopt the ResNet-200 \cite{He2016a} implemented in \cite{Gross2016}. (ii) \textbf{Going wider} by increasing the bottleneck width. (iii) \textbf{Increasing cardinality} by doubling $C$.

Table~\ref{tab:2x-complexity} shows that increasing complexity by 2\m~ consistently reduces error \vs the ResNet-101 baseline (22.0\%). But the improvement is small when going deeper (ResNet-200, by 0.3\%) or wider (wider ResNet-101, by 0.7\%).

On the contrary, \emph{increasing cardinality $C$ shows much better results than going deeper or wider}. The 2\m64d ResNeXt-101 (\ie, doubling $C$ on 1\m64d ResNet-101 baseline and keeping the width) reduces the top-1 error by 1.3\% to 20.7\%. The 64\m4d ResNeXt-101 (\ie, doubling $C$ on 32\m4d ResNeXt-101 and keeping the width) reduces the top-1 error to 20.4\%.

We also note that 32\m4d ResNet-101 (21.2\%) performs better than the deeper ResNet-200 and the wider ResNet-101, even though it has only $\sim$50\% complexity. This again shows that cardinality is a more effective dimension than the dimensions of depth and width.

\vspace{.5em}
\noindent\textbf{Residual connections.}
The following table shows the effects of the residual (shortcut) connections:
{
\renewcommand\arraystretch{1.05}
\begin{center}
\small
\begin{tabular}{l|c|c|c}
\hline
	& \footnotesize {setting}
	& \footnotesize  {w/ residual}
	& \footnotesize  {w/o residual} \\
\hline
ResNet-50		& 1 \m~64d		& 23.9 & 31.2 \\
ResNeXt-50		& 32 \m~4d		& \textbf{22.2} & 26.1 \\
\hline
\end{tabular}
\end{center}
}
\noindent
Removing shortcuts from the ResNeXt-50 increases the error by 3.9 points to 26.1\%. Removing shortcuts from its ResNet-50 counterpart is much worse (31.2\%). These comparisons suggest that the residual connections are helpful for \emph{optimization},
whereas aggregated transformations are stronger \emph{representations}, as shown by the fact that they perform consistently better than their counterparts with or without residual connections.

\vspace{.5em}
\noindent\textbf{Performance.}
For simplicity we use Torch's built-in grouped convolution implementation, without special optimization.
We note that this implementation was brute-force and not parallelization-friendly.
On 8 GPUs of NVIDIA M40, training 32\m4d ResNeXt-101 in Table~\ref{tab:preserve-comp} takes 0.95s per mini-batch, \vs 0.70s of ResNet-101 baseline that has similar FLOPs. We argue that this is a reasonable overhead. We expect carefully engineered lower-level implementation (\eg, in CUDA) will reduce this overhead. We also expect that the inference time on CPUs will present less overhead. Training the 2\m complexity model (64\m4d ResNeXt-101) takes 1.7s per mini-batch and 10 days total on 8 GPUs.

\vspace{.5em}
\noindent\textbf{Comparisons with state-of-the-art results.}
Table~\ref{tab:more} shows more results of single-crop testing on the ImageNet validation set. In addition to testing a 224$\times$224 crop, we also evaluate a 320$\times$320 crop following \cite{He2016a}. Our results compare favorably with ResNet, Inception-v3/v4, and Inception-ResNet-v2, achieving a single-crop top-5 error rate of 4.4\%. In addition, our architecture design is much simpler than all Inception models, and requires considerably fewer hyper-parameters to be set by hand.

ResNeXt is the foundation of our entries to the ILSVRC 2016 classification task, in which we achieved 2$^\text{nd}$ place. We note that many models (including ours) start to get saturated on this dataset after using multi-scale and/or multi-crop testing.
We had a single-model top-1/top-5 error rates of 17.7\%/3.7\% using the multi-scale dense testing in \cite{He2016}, on par with Inception-ResNet-v2's single-model results of 17.8\%/3.7\% that adopts multi-scale, multi-crop testing.
We had an ensemble result of 3.03\% top-5 error on the test set, on par with the winner's 2.99\% and Inception-v4/Inception-ResNet-v2's 3.08\% \cite{Szegedy2016}.

\setlength{\tabcolsep}{2pt}
\begin{table}[!htp]
\begin{center}
\small
\begin{tabular}{l|c|c|c|c}
\hline
& \multicolumn{2}{c|}{\footnotesize 224$\times$224}
& \multicolumn{2}{c}{\footnotesize 320$\times$320 / 299$\times$299} \\
\cline{2-5}
	& \footnotesize top-1 err
	& \footnotesize top-5 err
	& \footnotesize top-1 err
	& \footnotesize top-5 err \\
\hline
ResNet-101	\cite{He2016}							& 22.0 			& 6.0		& - 			& - \\
ResNet-200 \cite{He2016a}							& 21.7			& 5.8		& 20.1		& 4.8 \\
\hline
Inception-v3	 \cite{Szegedy2016a}				& -				& -			& 21.2 		& 5.6 \\
Inception-v4	 \cite{Szegedy2016}				& -				& -			& 20.0 		& 5.0 \\
Inception-ResNet-v2 \cite{Szegedy2016}	& -				& -			& 19.9 		& 4.9 \\
\hline
ResNeXt-101 (\textbf{64 \m~4d})				& {20.4}	& {5.3} 	& \textbf{19.1}			& \textbf{4.4} \\
\hline
\end{tabular}
\end{center}
\caption{State-of-the-art models on the ImageNet-1K validation set (single-crop testing).
The test size of ResNet/ResNeXt is 224$\times$224 and 320$\times$320 as in \cite{He2016a} and of the Inception models is 299$\times$299.}
\label{tab:more}
\end{table}

\begin{figure}[t]
\centering
\vspace{-1em}
\includegraphics[width=1.0\linewidth]{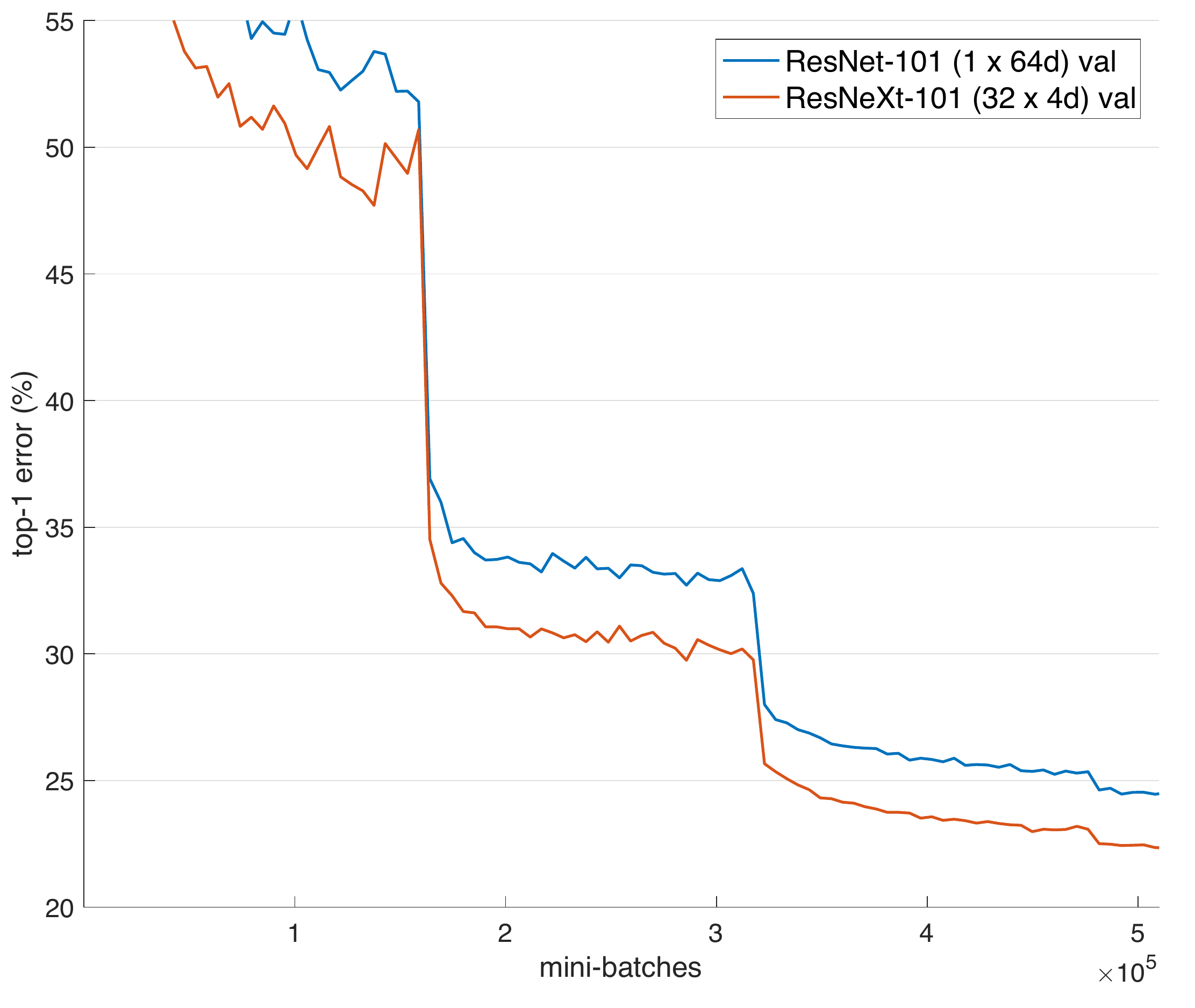}
\caption{\textbf{ImageNet-5K} experiments. Models are trained on the 5K set and evaluated on the original 1K validation set, plotted as a 1K-way classification task. ResNeXt and its ResNet counterpart have similar complexity.}
\label{fig:imagenet5k}
\end{figure}

\renewcommand\arraystretch{1.1}
\setlength{\tabcolsep}{2pt}
\begin{table}[t]
\begin{center}
\small
\begin{tabular}{l|c|x{32}|x{32}|x{32}|c}
\hline
	&
	& \multicolumn{2}{c|}{\footnotesize 5K-way classification}
	& \multicolumn{2}{c}{\footnotesize 1K-way classification} \\
\cline{3-6}
	& setting & \footnotesize top-1 & \footnotesize top-5
& \footnotesize top-1 & \footnotesize top-5 \\
\hline
ResNet-50		& 1 \m~64d		 & 45.5 & 19.4 & 27.1 		& 8.2\\
ResNeXt-50		& 32 \m~4d		 & \textbf{42.3} & \textbf{16.8} & \textbf{24.4} 	&	\textbf{6.6}  \\
\hline
ResNet-101		& 1 \m~64d		& 42.4 & 16.9 & 24.2		& 6.8   \\
ResNeXt-101	& 32 \m~4d		& \textbf{40.1} & \textbf{15.1} & \textbf{22.2} 	& \textbf{5.7}   \\
\hline
\end{tabular}
\end{center}
\caption{Error (\%) on \textbf{ImageNet-5K}. The models are trained on ImageNet-5K and tested \emph{on the ImageNet-1K val set}, treated as a 5K-way classification task or a 1K-way classification task at test time.
ResNeXt and its ResNet counterpart have similar complexity.
The error is evaluated on the single crop of 224$\times$224 pixels.}
\label{tab:imagenet5k}
\end{table}

\subsection{Experiments on ImageNet-5K}

The performance on ImageNet-1K appears to saturate. But we argue that this is not because of the capability of the models but because of the complexity of the dataset. Next we evaluate our models on a larger ImageNet subset that has 5000 categories.

Our 5K dataset is a subset of the full ImageNet-22K set \cite{Russakovsky2015}. The 5000 categories consist of the original ImageNet-1K categories and additional 4000 categories that have the largest number of images in the full ImageNet set.
The 5K set has 6.8 million images, about 5$\times$ of the 1K set.
There is no official train/val split available, so we opt to evaluate on the original ImageNet-1K validation set.
On this 1K-class val set, the models can be evaluated as a 5K-way classification task (all labels predicted to be the other 4K classes are automatically erroneous) or as a 1K-way classification task (softmax is applied only on the 1K classes) at test time.

The implementation details are the same as in Sec.~\ref{sec:impl}. The 5K-training models are all trained from scratch, and are trained for the same number of mini-batches as the 1K-training models (so 1/5$\times$ epochs).
Table~\ref{tab:imagenet5k} and Fig.~\ref{fig:imagenet5k} show the comparisons under preserved complexity. ResNeXt-50 reduces the 5K-way top-1 error by \textbf{3.2\%} comparing with ResNet-50, and ResNetXt-101 reduces the 5K-way top-1 error by \textbf{2.3\%} comparing with ResNet-101.
Similar gaps are observed on the 1K-way error.
These demonstrate the stronger representational power of ResNeXt.

Moreover, we find that the models trained on the 5K set (with 1K-way error 22.2\%/5.7\% in Table~\ref{tab:imagenet5k}) perform competitively comparing with those trained on the 1K set (21.2\%/5.6\% in Table~\ref{tab:preserve-comp}), evaluated on the same 1K-way classification task on the validation set. This result is achieved without increasing the training time (due to the same number of mini-batches) and without fine-tuning. We argue that this is a promising result, given that the training task of classifying 5K categories is a more challenging one.

\subsection{Experiments on CIFAR}

\newcommand{\blockc}[2]{\fontsize{6pt}{.8em}\selectfont \(\left[\begin{array}{c}\text{1$\times$1, #2}\\[-.1em] \text{3$\times$3, #2}\\[-.1em] \text{1$\times$1, #1}\end{array}\right]\)
}

We conduct more experiments on CIFAR-10 and 100 datasets \cite{Krizhevsky2009}.
We use the architectures as in \cite{He2016} and replace the basic residual block by the bottleneck template of {\fontsize{6pt}{.8em}\selectfont \(\left[\begin{array}{c}\text{1$\times$1, 64}\\[-.1em] \text{3$\times$3, 64}\\[-.1em] \text{1$\times$1, 256}\end{array}\right]\)}.
Our networks start with a single 3$\times$3 conv layer, followed by 3 stages each having 3 residual blocks, and end with average pooling and a fully-connected classifier (total 29-layer deep), following \cite{He2016}.
We adopt the same translation and flipping data augmentation as \cite{He2016}.
Implementation details are in the appendix.

\begin{figure}[t]
\centering
\vspace{-1em}
\includegraphics[width=1.0\linewidth]{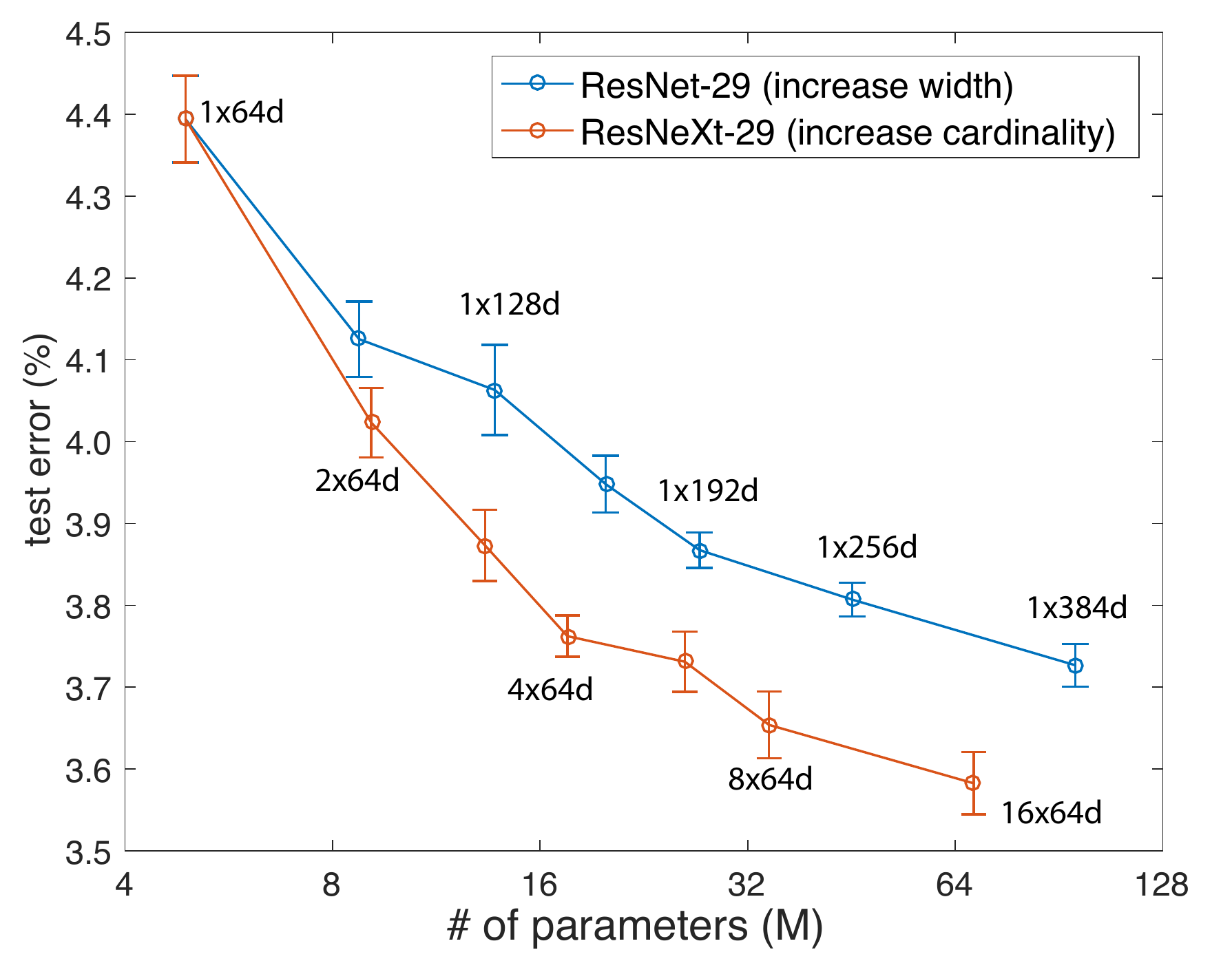}
\caption{Test error \vs model size on CIFAR-10.
The results are computed with 10 runs, shown with standard error bars. The labels show the settings of the templates.}
\label{fig:cifar}
\end{figure}

We compare two cases of increasing complexity based on the above baseline: (i) \emph{increase cardinality} and fix all widths, or (ii) \emph{increase width} of the bottleneck and fix cardinality $=1$.
We train and evaluate a series of networks under these changes.
Fig.~\ref{fig:cifar} shows the comparisons of test error rates \vs model sizes.
We find that increasing cardinality is more effective than increasing width, consistent to what we have observed on ImageNet-1K.
Table~\ref{tab:cifar10} shows the results and model sizes, comparing with the Wide ResNet \cite{Zagoruyko2016a} which is the best published record.
Our model with a similar model size (34.4M) shows results better than Wide ResNet.
Our larger method achieves 3.58\% test error (average of 10 runs)
% \footnote{The best single run has an error rate of 3.38\%.}
on CIFAR-10 and 17.31\% on CIFAR-100. To the best of our knowledge, these are the state-of-the-art results (with similar data augmentation) in the literature including unpublished technical reports.

\renewcommand\arraystretch{1.05}
\setlength{\tabcolsep}{4pt}
\begin{table}[t]
\begin{center}
\small
\begin{tabular}{l|c|c|c}
\hline
	& \footnotesize \# params
	& \footnotesize  CIFAR-10
	& \footnotesize  CIFAR-100 \\
\hline
Wide ResNet \cite{Zagoruyko2016a}		&  36.5M & 4.17 & 20.50 \\
\hline
ResNeXt-29, 8\m64d	&  34.4M &  3.65 	& 17.77 \\
ResNeXt-29, 16\m64d	&  68.1M & \textbf{3.58} & \textbf{17.31} \\
\hline
\end{tabular}
\end{center}
\caption{Test error (\%) and model size on CIFAR. Our results are the average of 10 runs.}
\label{tab:cifar10}
\end{table}

\renewcommand\arraystretch{1.05}
\setlength{\tabcolsep}{8pt}
\begin{table}[t]
\begin{center}
\small
\begin{tabular}{l|c|c|c}
\hline
	& \footnotesize \tabincell{c}{setting}
	& \footnotesize  \tabincell{c}{AP@0.5}
	& \footnotesize  \tabincell{c}{AP} \\
\hline
ResNet-50		& 1 \m~64d		& 47.6 & 26.5 \\
ResNeXt-50		& 32 \m~4d		& \textbf{49.7} & \textbf{27.5} \\
\hline
ResNet-101		& 1 \m~64d		& 51.1 & 29.8 \\
ResNeXt-101	& 32 \m~4d		& \textbf{51.9} & \textbf{30.0} \\
\hline
\end{tabular}
\end{center}
\caption{Object detection results on the COCO \texttt{minival} set. ResNeXt and its ResNet counterpart have similar complexity.}
\label{tab:det}
\end{table}

\subsection{Experiments on COCO object detection}

Next we evaluate the generalizability on the COCO object detection set \cite{Lin2014}. We train the models on the 80k training set plus a 35k val subset and evaluate on a 5k val subset (called \texttt{minival}), following \cite{Bell2016}. We evaluate the COCO-style Average Precision (AP) as well as AP@IoU=0.5 \cite{Lin2014}.
We adopt the basic Faster R-CNN \cite{Ren2015} and follow \cite{He2016} to plug ResNet/ResNeXt into it.
The models are pre-trained on ImageNet-1K and fine-tuned on the detection set.
Implementation details are in the appendix.

Table~\ref{tab:det} shows the comparisons. On the 50-layer baseline, ResNeXt improves AP@0.5 by 2.1\% and AP by 1.0\%, without increasing complexity. ResNeXt shows smaller improvements on the 101-layer baseline. We conjecture that more training data will lead to a larger gap, as observed on the ImageNet-5K set.

It is also worth noting that recently ResNeXt has been adopted in Mask R-CNN \cite{he2017mask} that achieves state-of-the-art results on COCO instance segmentation and object detection tasks.

\subsection*{Acknowledgment} S.X. and Z.T.'s research was partly supported by NSF IIS-1618477. The authors would like to thank Tsung-Yi Lin and Priya Goyal for valuable discussions.

\appendix

\section{Implementation Details: CIFAR}
\label{app:cifar}

We train the models on the 50k training set and evaluate on the 10k test set.
The input image is 32$\times$32 randomly cropped from a zero-padded 40$\times$40 image or its flipping, following \cite{He2016}.
No other data augmentation is used.
The first layer is 3$\times$3 conv with 64 filters.
There are 3 stages each having 3 residual blocks, and the output map size is 32, 16, and 8 for each stage \cite{He2016}.
The network ends with a global average pooling and a fully-connected layer.
Width is increased by 2$\times$ when the stage changes (downsampling), as in Sec.~\ref{sec:template}.
%We adopt the pre-activation style block as in \cite{He2016a}.
The models are trained on 8 GPUs with a mini-batch size of 128, with a weight decay of 0.0005 and a momentum of 0.9.
We start with a learning rate of 0.1 and train the models for 300 epochs, reducing the learning rate at the 150-th and 225-th epoch.
Other implementation details are as in \cite{Gross2016}.

\section{Implementation Details: Object Detection}
\label{app:det}

We adopt the Faster R-CNN system \cite{Ren2015}.
For simplicity we do not share the features between RPN and Fast R-CNN.
In the RPN step, we train on 8 GPUs with each GPU holding 2 images per mini-batch and 256 anchors per image.
We train the RPN step for 120k mini-batches at a learning rate of 0.02 and next 60k at 0.002.
In the Fast R-CNN step, we train on 8 GPUs with each GPU holding 1 image and 64 regions per mini-batch. 
We train the Fast R-CNN step for 120k mini-batches at a learning rate of 0.005 and next 60k at 0.0005,
We use a weight decay of 0.0001 and a momentum of 0.9. 
Other implementation details are as in \url{https://github.com/rbgirshick/py-faster-rcnn}.

{%\small
%\fontsize{9.5pt}{0.6em}\selectfont
\bibliographystyle{ieee}
\bibliography{resnext}
}

\end{document}